\documentclass[letterpaper]{article} 
\usepackage[preprint]{aaai2027}  
\usepackage[hyphens]{url}  
\usepackage{graphicx} 
\usepackage{natbib}  
\usepackage{caption} 
\usepackage{amsmath}
\usepackage{amssymb}
\usepackage{booktabs}

\title{MedCache: Efficient and Temporally Valid Memory for Longitudinal Clinical Agents}
\author{
    Hei Ting (Una) Chan\equalcontrib,
    Chenwei Wu\equalcontrib,
    Xueshen Liu,
    Boyuan Zheng,\\
    Liyue Shen,
    Jiasi Chen,
    Z. Morley Mao
}
\affiliations{
    University of Michigan, Ann Arbor, MI, USA\\
    \{hei, chenweiw, liuxs, boyuann, liyues, jiasi, zmao\}@umich.edu
}

\begin{document}

\maketitle

\begin{abstract}
Longitudinal clinical agents must maintain an evolving patient state from evidence distributed across visits, time points, and specialties. However, how agent memory should be designed for this setting remains unclear. We introduce a benchmark of multi-visit, multi-specialty patient records that evaluates long-context evidence retrieval, cross-time evidence aggregation, and cross-specialty clinical reasoning. Using this benchmark, we systematically study four memory design choices: curation, organization, retrieval, and memory-augmented reasoning. We find that temporal validity is more important than simply retaining more history; specialty-factorized memory reduces context but can hide shared evidence; and multiple agents help when specialists must reason together, not merely when evidence comes from multiple memories. Guided by these findings, we propose \textit{MedCache}, a hybrid framework that constructs temporally valid patient memory, organizes evidence into overlapping specialty views, routes each query to relevant memories, and adaptively invokes one or multiple specialists. Experiments show that MedCache improves reasoning accuracy and memory efficiency over strong single-agent and multi-agent baselines, while generalizing across model backbones and external datasets.

\end{abstract}


\section{Introduction}
Personalized medical AI agents are rapidly moving beyond single-shot question-answering prototypes toward long-term care systems for continuous patient support and longitudinal disease management~\cite{antgroup2025afu,uhs2025hippocratic,shi2026medxiaohe}. In real-world deployment, agents must maintain an evolving representation of patient state from information distributed across visits, specialties, and care settings. Their decisions may depend on long-range clinical history, including medication changes and physiological trajectories, while forgetting patient-specific constraints or relying on outdated information can lead to unsafe recommendations. 
Reliable, longitudinal agent memory is therefore a foundational requirement for safe, personalized, and continuous clinical reasoning.

Agent-memory research has made notable progress in general domains beyond medical applications~\cite{zhang2025survey}, such as long conversations and extended coding sessions, through both single-agent and multi-agent approaches. 
Single-agent systems (SAS) leverage approaches like retrieval-augmented generation~\cite{arslan2024survey}, summarization~\cite{yu2025memagent}, or structured memory~\cite{chhikara2025mem0} to give one agent unified access to past information. Multi-agent systems (MAS) instead divide memory and reasoning across role-specialized agents, allowing each agent to focus on a smaller and more relevant context~\cite{zhang2026g,wu2026memory}. 
Longitudinal medical records, however, introduce additional challenges. Laboratory values, diagnoses, and medications may change over time, while clinically relevant evidence may be spread across departments and recorded months or years apart. 
A single-agent system can preserve a holistic view of the patient, but it may need to process a large amount of outdated, irrelevant, and clinically diverse information in one context. 
A multi-agent system may better match the multi-specialty structure of patient history, but how to efficiently organize this long-context memory with MAS is nontrivial: it can split shared evidence across agents, repeat the same information across multiple calls, or produce conflicting specialist recommendations. 
An important research question remains understudied: \textit{how to design effective and efficient agent-memory mechanisms for longitudinal medical care?}

To better understand existing memory approaches’ advantages and limitations in longitudinal medical care, we present \textbf{LEMMA-Bench} (Longitudinal EHR Memory for Medical Agents Benchmark), the first benchmark to evaluate memory designs for both SAS and MAS in longitudinal medical care settings. This benchmark, curated from de-identified longitudinal patient records,  targets  \textit{multi-visit, multiple EHR data types and multi-department} settings in real-world clinical practice. The benchmark evaluates long-context evidence retrieval, cross-time evidence aggregation, and cross-specialty clinical reasoning. We compare strong SAS and MAS baselines while systematically analyzing four dimensions of agent memory: \emph{memory curation}, including how evolving facts are updated and temporally validated; \emph{memory organization}, including shared and specialty-distributed storage; \emph{memory retrieval}, including how agents retrieve different memory segments; and \emph{memory-augmented reasoning}, including single-agent and coordinated specialist-based multi-agent reasoning. 

Our study yields several critical findings. First, retaining more history does not guarantee better memory curation: longitudinal reasoning requires curating a valid patient state, as raw long-context and semantic retrieval can still contain outdated or superseded evidence that confuses agents. Second, distributing memory into focused clinical specialist views substantially reduces the context processed per query, but strict specialty boundaries can hide evidence relevant to multiple domains. Finally, memory-augmented MAS reasoning is not uniformly better than SAS despite increased inference costs. A single agent can usually integrate evidence retrieved from several memories, while multiple specialists are more helpful in the scenario when distinct clinical constraints must be evaluated and reconciled. 
Together, these findings show that no fixed memory or agent configuration in existing works is consistently optimal across varying longitudinal clinical queries. 

Motivated by these findings, we propose \textit{MedCache}, a hybrid framework that curates temporally valid patient memory, organizes evidence into overlapping specialty views, and routes each query to the relevant memories. MedCache then adaptively invokes either a single reader or multiple coordinated specialists according to the reasoning requirements of the query. Extensive evaluation shows that MedCache improves reasoning accuracy and memory efficiency and generalization.

Our main contributions are threefold:
\begin{itemize}
\item We introduce a benchmark for longitudinal clinical agent memory using multi-visit, multi-specialty patient records, covering long-context evidence retrieval, cross-time evidence aggregation, and cross-specialty clinical reasoning.

\item We systematically study four key memory design choices---\emph{memory curation}, \emph{memory organization}, \emph{memory retrieval}, and \emph{memory-augmented reasoning} and derive useful principles for effective longitudinal clinical memory.

\item Guided by these findings, we propose \textit{MedCache}, a hybrid framework that combines temporally valid memory, overlapping specialty views, routed access, and adaptive specialist reasoning. Extensive experiments demonstrate improved accuracy and memory efficiency.

\end{itemize}

\section{Related Work}
\label{sec:related-work}

\paragraph{Longitudinal clinical benchmarks.} LongHealth evaluates extraction, negation, and temporal sorting over extended synthetic patient cases \cite{adams2024longhealth}, while EHRNoteQA includes clinician-reviewed questions that may require evidence across multiple discharge summaries \cite{kweon2024ehrnoteqa}. Interactive benchmarks extend this setting to agent actions: MedAgentBench provides a FHIR-compliant EHR environment \cite{jiang2025medagentbench}, and MedicalAgentsBench evaluates complex diagnosis and treatment-planning problems under accuracy and cost constraints \cite{tang2025medagentsbench}. LEMMA-Bench targets two axes that govern longitudinal clinical state: temporal validity and overlapping specialty ownership, and separately measures state maintenance, owner selection, and cross-specialty reasoning.

\paragraph{Long-context and memory.} Long-context models remain sensitive to evidence position \cite{liu2024lost}, and medical RAG improves context selection through retrieval \cite{xiong2024medrag}. Persistent-memory systems manage extended interactions through virtual context tiers \cite{packer2023memgpt}, consolidated salient memories \cite{chhikara2025mem0}, or temporally aware knowledge graphs \cite{rasmussen2025zep}. MedCache specializes these ideas for longitudinal care: provenance-linked clinical facts preserve bounded trajectories, temporal validation determines which facts enter memory, and specialty ownership determines where shared evidence resides. Query relevance then selects among valid, owner-specific memories.

\paragraph{Medical multi-agent reasoning and serving.} MedAgents organizes multidisciplinary consultation through iterative consensus \cite{tang2024medagents}, while MDAgents adapts solo or group collaboration to case complexity \cite{kim2024mdagents}. Med-SRAF adds semantic routing and evidence fusion \cite{li2026medsraf}, and ClinicalAgents combines mutable working memory with search-based orchestration \cite{ge2026clinicalagents}. MedCache moves specialty factorization into persistent patient memory and uses a learned owner gate to select the compact evidence-bearing team for each query. Locally answerable queries use one specialist after cross-owner evidence delivery; coupled decisions trigger coordinated execution. At the serving layer, PagedAttention shares KV-cache blocks \cite{kwon2023vllm} and SGLang reuses common prefixes through RadixAttention \cite{zheng2024sglang}. MedCache aligns this reuse model with clinical ownership: stable specialty guidelines form reusable prefixes, and routed patient evidence forms a compact dynamic suffix.

\section{Benchmarking and Analyzing Longitudinal Medical Agent Memory Failures}
This section details the data collection, processing and technical implementation processes for building our benchmark for longitudinal medical care, for assessing how agent-memory systems maintain and use longitudinal patient information for varying clinical queries. 
LEMMA-Bench aims to address three research questions:

\begin{itemize}
\item \textbf{RQ1:} How well do existing SAS and MAS agent-memory methods perform on longitudinal clinical reasoning?

\item \textbf{RQ2:} Which design choices in memory curation, organization, retrieval, and reasoning explain the memory system failures?

\item \textbf{RQ3:} Can these findings guide the design of effective and efficient agent memory mechanisms for longitudinal clinical care?

\end{itemize}
\label{sec:primary-benchmark}
\subsection{Benchmark Construction and Evaluation} LEMMA-Bench evaluates three core agent memory capabilities: 1) Long-context evidence retrieval, where agents must recover a patient-specific fact from a long history containing recurring measurements, unrelated events, and outdated observations. 2) Cross-time evidence aggregation, where agents must combine information from multiple visits or time points. These questions may require identifying a physiological trend, tracking a medication from initiation to discontinuation, or determining whether a previously documented condition remains active. 3) Cross-specialty clinical reasoning, where agents must integrate patient evidence and clinical criteria associated with different specialties. 

The benchmark is constructed from de-identified longitudinal records in MIMIC-IV v3.1~\cite{johnson2023mimic}. We select 91 patients with long patient histories, yielding 1,152 admissions and an average record span of 5.8 years. The resulting histories contain recurring laboratory measurements, prescriptions, discharge summaries, and radiology reports, creating realistic temporal and cross-specialty memory challenges. Questions are constructed from these evidence units and relevant clinical guidelines, and each question is labeled with the exact record entries needed to answer it. Across these histories the benchmark covers 193,394 laboratory measurements, a mean of 2,125 per patient and a median of 1,324, drawn from 381 distinct laboratory concepts of which an average patient has 88. Evidence is owned by a registry of ten specialties, and clinical criteria are supplied by real-world clinical guidelines~\cite{khwaja2012kdigo,yancy20172017,elsayed2025introduction}. Full benchmark dataset details can be found in the Appendix.
\begin{figure}
    \centering
    \includegraphics[width=\columnwidth]{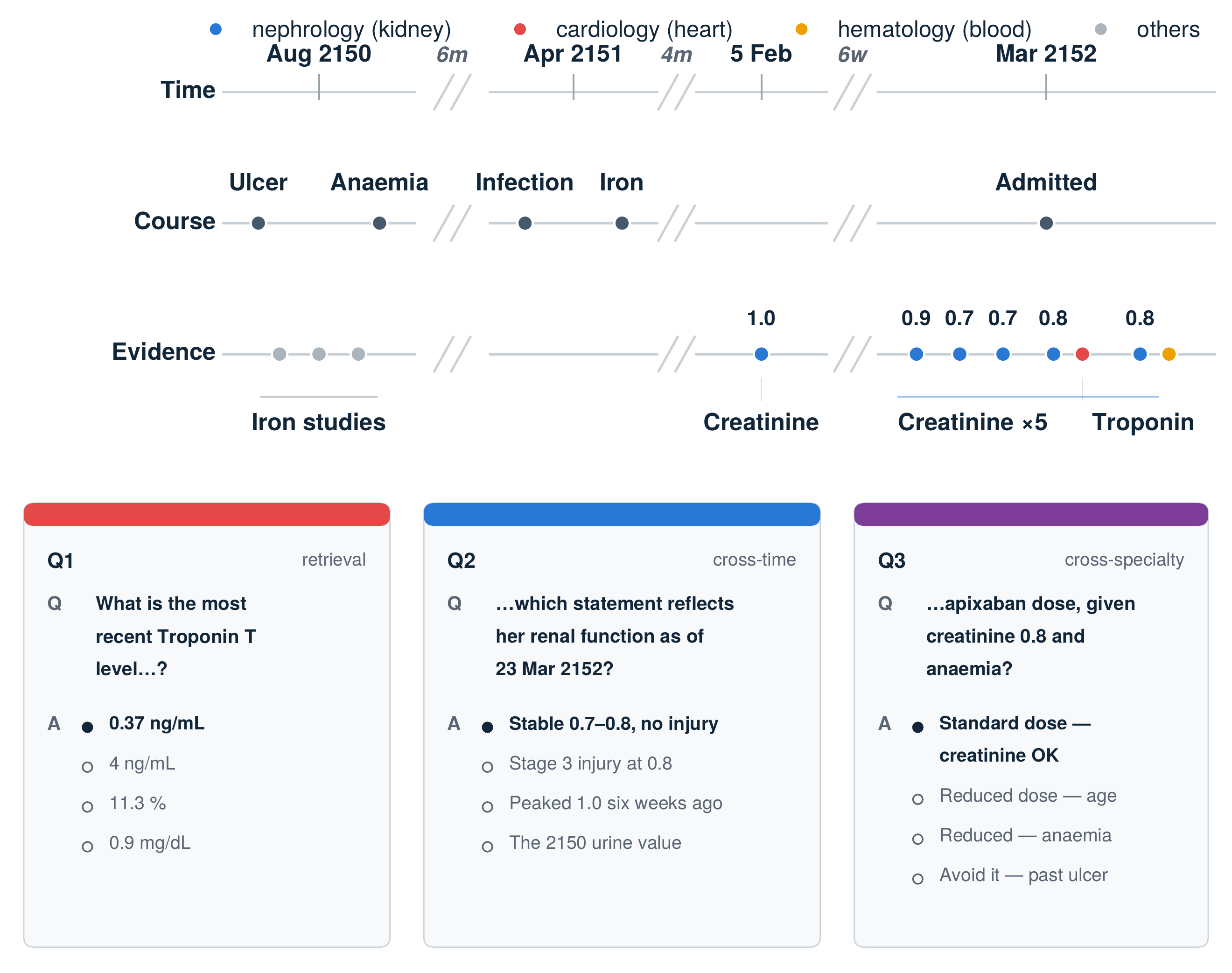}
    \caption{An example benchmark patient with question examples.}
    \label{fig:bench_example}
\end{figure}
Human experts manually draft the questions and the initial pass yielded 474 candidate question-answer pairs. Each generated question first undergoes an automated filtering process to ensure that all questions are answerable with labeled evidence and not guessable without any patient records. Filtered candidates additionally go through two human expert checks to audit gold answer correctness, option ambiguity, evidence sufficiency, task-type label agreement and temporal validity of the question. The final benchmark contains 82 long-context evidence retrieval questions, 46
cross-time evidence aggregation questions, and 60 cross-specialty reasoning questions, all in multiple-choice format. Fig.~\ref{fig:bench_example} shows a patient example.

We run experiments on 2 open-source LLM backbones, MoE Qwen 3.6-35B-A3B~\cite{qwen36_35b_a3b} model and dense Mistral-24B model served through vLLM with automatic prefix caching. 
We evaluate SAS memory systems including raw long-context baselines, RAG~\cite{lewis2020retrieval}, Zep~\cite{rasmussen2025zep}, MemGPT(Letta)~\cite{packer2023memgpt}, Mem0~\cite{chhikara2025mem0}, and MAS memory systems including MDAgents~\cite{kim2024mdagents}, Med-SRAF~\cite{li2026medsraf}, ClinicalAgents~\cite{ge2026clinicalagents}. We analyze four core dimensions of an agent-memory system: 
1) \textit{Memory curation}: how evolving diagnoses, medications, measurements, and other patient facts are retained and updated over time;
2) \textit{Memory organization}: whether patient memory is centralized (SAS), shared (MAS), or divided into specialty-focused views (MAS) ; 
3) \textit{Memory retrieval}: which parts of the patient memory are selected for each query; 
and 4) \textit{Memory-augmented reasoning}: whether the selected evidence is processed by only one agent or coordinated specialists. 
For quantitative evaluation, we mainly report the overall and category-wise multiple-choice question accuracy. The model's memory formation and answer generation processes were stored for qualitative analysis. 
\begin{table}[t]
\centering
\caption{Performance of existing memory systems on Qwen 3.6-35B-A3B. Input tokens for MAS are summed across agents. Evid., C-Time, and C-Spec. denote accuracy on the long-context evidence-retrieval, cross-time aggregation, and cross-specialty reasoning questions, respectively.}
\label{tab:existing_memory_results}
\scriptsize
\setlength{\tabcolsep}{4.6pt}
\renewcommand{\arraystretch}{1.08}
\begin{tabular}{@{}lcccccc@{}}
\toprule
System &
Overall [95\% CI] &
Evid. &
C-Time &
C-Spec. &
In tok. &
Calls \\
\midrule

\multicolumn{7}{@{}l}{\textit{Raw context}} \\
Long-Context
& .705 [.651,.756]
& .735 & .667 & .707
& 29.7k & 1.00 \\
\midrule

\multicolumn{7}{@{}l}{\textit{Single-agent memory}} \\
RAG
& .489 [.412,.561]
& .294 & .372 & .850
& 14.5k & 1.00 \\

Zep
& .731 [.691,.784]
& .886 & .870 & .413
& 3.9k & 1.00 \\

MemGPT
& .611 [.589,.699]
& .727 & .750 & .347
& 4.1k & 1.00 \\

Mem0
& .381 [.329,.433]
& .417 & .361 & .347
& 1.6k & 1.00 \\
\midrule

\multicolumn{7}{@{}l}{\textit{Multi-agent memory}} \\
MDAgents
& .862 [.796,.923]
& .859 & .953 & .800
& 23.9k & 2.69 \\

Med-SRAF
& .787 [.724,.846]
& .659 & .791 & \textbf{.967}
& 8.8k & 4.93 \\

ClinicalAgents
& \textbf{.915 [.870,.956]}
& \textbf{.941} & \textbf{.977} & .833
& 31.5k & 2.00 \\
\bottomrule
\end{tabular}
\end{table}
\subsection{Benchmark Results and Memory Failure Analysis}

\subsubsection{RQ1: How Well Do Existing Memory Systems Perform?}
\label{sec:rq1}

Tab.~\ref{tab:existing_memory_results} compares representative raw-context, single-agent memory, and multi-agent memory systems across long-context retrieval, cross-time aggregation, and cross-specialty reasoning. No method performs consistently best across all three capabilities. Among single-agent approaches, Zep achieves the strongest overall accuracy at .731 while using only 3.9k input tokens, with strong retrieval (.886) and cross-time performance (.870) but much lower cross-specialty accuracy (.413). MemGPT shows a similar but weaker pattern, while Mem0 performs poorly across all tasks. RAG exhibits the opposite imbalance, reaching .850 on cross-specialty questions but only .294 on retrieval and .372 on cross-time aggregation. The raw long-context baseline also remains limited at .705 overall despite processing 29.7k input tokens.

Multi-agent systems generally achieve higher overall accuracy, but their gains likewise vary by task and cost. ClinicalAgents obtains the strongest overall result at .915 and performs best on retrieval (.941) and cross-time aggregation (.977), but reaches only .833 on cross-specialty reasoning. Med-SRAF instead achieves the highest cross-specialty accuracy (.967), while performing substantially worse on retrieval (.659) and cross-time aggregation (.791). MDAgents provides a more balanced result, but requires 23.9k input tokens and 2.69 model calls per question. Overall, these results do not identify a uniformly superior single-agent or multi-agent architecture in terms of effectiveness and efficiency.
\subsubsection{RQ2: Which design choices in memory curation, organization, retrieval, and reasoning explain the memory system failures?}

\begin{table}[t]
\centering
\caption{Normalized error composition by system. Each row sums to one.}
\label{tab:normalized_error_composition}
\small
\setlength{\tabcolsep}{6.5pt}
\begin{tabular}{@{}lccc@{}}
\toprule
System & C1: Outdated & C2: Missed & C3: Fabricated \\
\midrule
Long-Context          & 0.549 & 0.413 & 0.038 \\
RAG                   & 0.536 & 0.440 & 0.025 \\
Mem0                  & 0.599 & 0.284 & 0.117 \\
MemGPT               & 0.407 & 0.479 & 0.114 \\
Zep                   & 0.721 & 0.210 & 0.069 \\
MDAgents              & 0.797 & 0.166 & 0.037 \\
Med-SRAF              & 0.639 & 0.312 & 0.048 \\
ClinicalAgents  & 0.572 & 0.373 & 0.056 \\
\bottomrule
\end{tabular}
\end{table}

\paragraph{Memory curation.}
To understand the failures of memory curation, Tab.~\ref{tab:normalized_error_composition} qualitatively categorizes memory systems' errors into writing outdated evidence into memory, failure to recover required evidence, and reasoning fabrication. Results show that \textbf{outdated evidence in the curation stage constitutes the largest share of observed errors for nearly every system}, accounting for 54.9\% of Long-Context errors, 72.1\% of Zep errors, and 79.7\% of MDAgents errors. Fabrication is less common overall, but remains non-negligible for compressed-memory systems such as Mem0 and MemGPT, where it represents 11.7\% and 11.4\% of errors, respectively. Our qualitative analysis identifies that a representative stale-evidence failure occurs when the system is asked for a patient’s most recent partial thromboplastin time (PTT). The correct value is 38.6 seconds, yet Long-Context, Mem0, and Zep all return 102.8 seconds. This value is not fabricated but an earlier PTT from the same admission, indicating that memory systems identify a plausible clinical fact but fail to determine that it has been superseded. Therefore the dominant longitudinal-memory failure is an inability to determine which facts remain current.

\begin{table}[t]
\centering
\caption{Ablation over memory organization, agent activation, and
output integration. }
\label{tab:memory-organization}
\scriptsize
\setlength{\tabcolsep}{2.3pt}
\renewcommand{\arraystretch}{1.05}
\begin{tabular}{@{}lllcccc@{}}
\toprule
Mem. & Act. & Output
& Overall & Evid. & C-Time & C-Spec. \\
\midrule
Centralized
& Single
& Reader (Zep)
& .731
& .886
& .870
& .413 \\
\midrule
Shared
& All
& Direct
& \textbf{.920}
& .927
& .870
& .950 \\
Shared
& All
& Coord.
& .910
& .915
& .848
& .950 \\
Shared
& Routed
& Direct (MDAgents)
& .862
& .859
& .953
& .800 \\
Shared
& Routed
& Coord. (ClinicalAgents)
& .915
& \textbf{.941}
& \textbf{.977}
& .833 \\
\midrule
Excl.
& Single
& Reader
& .734
& .720
& .674
& .800 \\
Excl.
& All
& Direct (Med-SRAF)
& .787
& .659
& .791
& \textbf{.967} \\
Excl.
& All
& Coord.
& .766
& .720
& .630
& .933 \\
Excl.
& Routed
& Direct
& .750
& .683
& .652
& .917 \\
Excl.
& Routed
& Coord.
& .750
& .683
& .674
& .900 \\
\midrule
Non-excl.
& Single
& Reader
& .846
& .927
& .891
& .700 \\
Non-excl.
& All
& Direct
& \textbf{.920}
& .915
& .891
& .950 \\
Non-excl.
& All
& Coord.
& .915
& .927
& .870
& .933 \\
Non-excl.
& Routed
& Direct
& \textbf{.920}
& .927
& .913
& .917 \\
Non-excl.
& Routed
& Coord.
& .894
& .927
& .870
& .867 \\
\bottomrule
\end{tabular}
\end{table}

\paragraph{Memory organization and memory-augmented reasoning.}
Existing agent-memory methods combine choices of memory organization and memory-augmented reasoning in different ways, making it difficult to determine which component drives their performance. To disentangle these effects, in Tab.~\ref{tab:memory-organization}, we construct controlled configurations that vary three dimensions while holding all other settings fixed. \emph{Memory organization} determines whether evidence is centralized, shared across agents, isolated in exclusive specialty partitions, or stored in non-exclusive partitions that permit cross-partition access. \emph{Agent activation} determines whether one reader, all specialists, or only routed specialists are invoked. \emph{Output integration} determines whether the system answers through one reader, direct specialist collaboration, or a separate coordinator. Parenthesized rows report complete published systems and provide reference points rather than controlled comparisons.

The controlled configurations show that \textbf{cross-partition accessibility is critical.} Exclusive partitions consistently reduce Evidence Retrieval and Cross-Time performance because relevant history may be stored outside the active specialist's memory. However, their strong Cross-Specialty results, including Med-SRAF's best score, show that specialization can remain effective even when longitudinal evidence access is limited. Among accessible-memory configurations, \textbf{the appropriate activation policy depends on the task.} Under shared memory, activating all specialists is stronger than routing, particularly for Cross-Specialty. Under non-exclusive memory, routed direct collaboration provides the strongest Cross-Time result, while all-specialist collaboration performs best on Cross-Specialty. A single reader remains competitive for Evidence and Cross-Time but is substantially weaker on Cross-Specialty. Finally, controlled comparisons show relatively close performance between a separate coordinator and direct collaboration.

\paragraph{Memory retrieval.}
In Tab.~\ref{tab:routing}, we compare six policies for selecting the specialty memories in MAS systems. \emph{Query BM25} and \emph{query dense} rank memories using lexical and embedding similarity to the question, respectively. \emph{Query+patient BM25} additionally incorporates the patient's current-state summary into the lexical query. \emph{LLM selection} asks the backbone model to identify the relevant evidence owners, whereas the \emph{learned gate} uses a lightweight classifier trained on query--owner pairs. \emph{Full memory} provides the complete patient memory without selection.

Lexical and dense retrieval identify the required specialties unreliably, with specialty recall between .688 and .727 and overall accuracy below .700. Adding patient context to BM25 provides only a small improvement, indicating that semantic relevance alone does not reliably reveal where the required longitudinal evidence is stored. Both the LLM selector and learned gate recover every required specialty and achieve .945 overall accuracy. The learned gate matches the LLM selector while reducing routing latency significantly. Our takeaway is that \textbf{effective memory retrieval depends on identifying the small set of specialties that own the required evidence, rather than retrieving memories solely by similarity or exposing agents to the complete patient history.}

\begin{table}[t]
\centering
\caption{Comparison of memory-retrieval policies.}
\label{tab:routing}
\scriptsize
\setlength{\tabcolsep}{3pt}
\renewcommand{\arraystretch}{1.08}
\begin{tabular}{@{}lccccc@{}}
\toprule
Retrieval policy &
Overall &
\shortstack{Specialty\\recall} &
\shortstack{Fraction of memory\\read} &
Context &
\shortstack{Route\\time} \\
\midrule
Query BM25
& .680 & .703 & .192 & 3{,}080 & 0\,ms \\
Query+patient BM25
& .688 & .727 & .203 & 3{,}146 & 0\,ms \\
Query dense
& .695 & .688 & .155 & 2{,}432 & 7\,ms \\
LLM selection
& \textbf{.945} & \textbf{1.000} & .170
& \textbf{1{,}581} & 343\,ms \\
\textbf{Learned gate}
& \textbf{.945} & \textbf{1.000} & .173
& 1{,}621 & \textbf{7\,ms} \\
Full memory
& .859 & \textbf{1.000} & 1.000
& 16{,}829 & 0\,ms \\
\bottomrule
\end{tabular}
\end{table}

\section{MedCache Design}
\begin{figure*}[t]
\centering
\includegraphics[width=\textwidth]{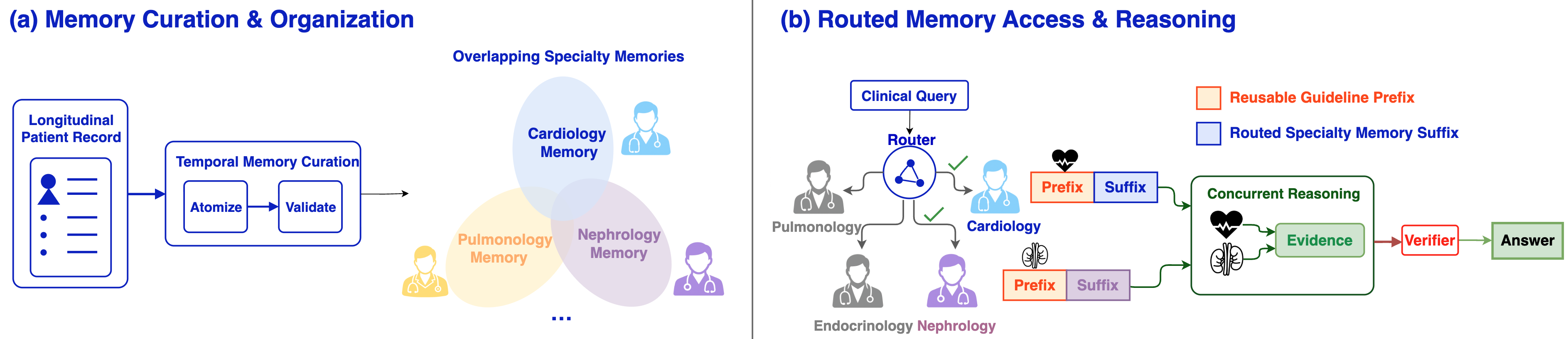}
\caption{Overview of the MedCache framework. 
(a) MedCache converts the longitudinal patient record into a temporally valid patient state and organizes valid facts into overlapping specialty memories. (b) At query time, the router selects the relevant specialty memories and guideline modules. Specialists reason over reusable guideline prefixes and routed patient memory suffixes. Coupled clinical decisions use concurrent specialist reasoning, coordination, and answer verification.
}
\label{fig:medcache_overview}
\end{figure*}

Inspired by the findings, we propose \textit{MedCache}, a hybrid agent-memory framework illustrated in Fig.~\ref{fig:medcache_overview}.
In Stage 1: temporal memory curation constructs a valid patient state from the longitudinal record, which is then organized into overlapping specialty memories, providing focused clinical views while preserving evidence shared across specialties. 
In Stage 2: a learned router selects the memories relevant to each query. An adaptive execution policy determines whether one or multiple specialists should reason over the selected evidence. Crucially, memory access and agent activation are separate decisions: a single specialist may integrate evidence retrieved from several specialty memories, while multiple specialists are activated only when the decision requires distinct clinical constraints to be evaluated and reconciled.

\paragraph{Temporal Memory Curation.}
As shown in Fig.~\ref{fig:medcache_overview}(a), MedCache converts structured and free-text EHR entries into normalized clinical facts. Each fact records its concept, value, unit, event and availability times, validity interval, evidence type, and source identifier. At query time, MedCache reconstructs the applicable patient state, including valid laboratory values, medication status, and changes in diagnoses and notes. For time-sensitive concepts, it preserves bounded trajectories within clinical windows, retaining the historical evidence needed for temporal reasoning.

\paragraph{Overlapping Specialty Memories.}
MedCache assigns each valid fact to one or more specialty memories through a concept-to-specialty registry. For example, creatinine belongs to nephrology, while potassium belongs to both nephrology and cardiology. Overlapping ownership keeps clinically shared evidence accessible across relevant domains without duplicating its underlying identity. Shared facts retain the same identifier across memories, which supports deduplication and consistent source citation when several memories are combined. MedCache also derives a high-recall, query-independent set of active specialties from the current patient state and refreshes it whenever the record changes. This set narrows the candidate evidence owners considered during query-time routing.

\paragraph{Routed Memory Access.}
As shown in Fig.~\ref{fig:medcache_overview}(b), a learned owner gate selects the specialty memories relevant to each query. The gate uses a frozen Qwen encoder and a two-layer classifier trained offline on query and owner pairs. For a patient evidence query, MedCache combines the requesting specialty memory with the predicted owner memory and removes duplicate facts. It ranks the remaining facts according to the query and selected guideline while preserving complete clinical trajectories. Each specialist receives a reusable, versioned guideline module as a stable prefix and the routed patient evidence as a dynamic suffix. vLLM reuses the cached model state when requests share the same guideline prefix.

\paragraph{Adaptive Specialist Reasoning.}
MedCache supports two reasoning paths. For an evidence-retrieval query, the owner gate selects the requesting specialty and the predicted evidence owner. MedCache deduplicates the union of their memory partitions and gives the resulting packet to a single reader, which produces the final answer directly. This path does not execute a specialist panel or call the coordinator. For a cross-specialty integration query, MedCache activates the patient-relevant specialists concurrently. Each specialist reads its own memory partition and guideline module and returns a structured assessment containing its decision, supporting evidence, uncertainty, and source identifiers. A coordinator then aligns shared evidence, reconciles the specialist assessments, and produces the final answer.

\paragraph{Answer Verification.}
Verification runs in two tiers. For criteria governed by a deterministic arithmetic or threshold rule in the specialist guidelines, MedCache recomputes the decision directly from the routed patient values and compares
it with the generated answer. Criteria with no such rule use a model-based verifier that checks whether the answer follows from the applicable guideline clause
and the cited evidence. Deterministic checks take precedence because they provide exact decisions for arithmetic and threshold-based criteria. Either tier records agreement or raises a
review flag for re-execution, and neither rewrites the answer.

MedCache separates system work into offline preparation, patient updates, and query-time execution. Offline preparation defines specialty ownership, versions guideline modules, and trains the owner gate. Patient updates refresh clinical facts, temporal validity, trajectories, specialty memories, and the active specialty set. Query-time execution covers routing, evidence assembly, specialist reasoning, coordination, and verification.
\begin{figure}
    \centering
    \includegraphics[width=\columnwidth]{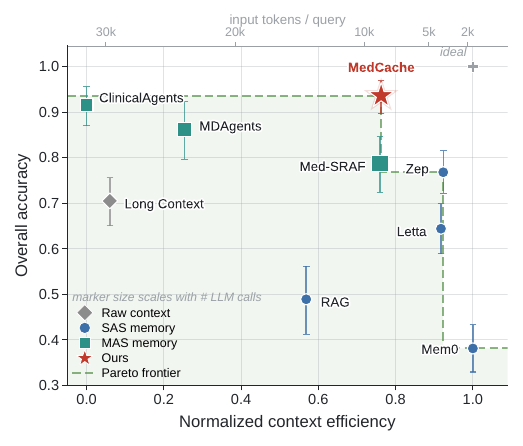}
    \caption{Accuracy--context trade-off on LEMMA-Bench. Higher normalized
    context efficiency indicates fewer input tokens.}
    \label{fig:accuracy-context}
\end{figure}
\section{Analysis of MedCache Results}
\label{sec:experiments}

In this section, we address \textbf{RQ3: Can these findings guide the design of effective and efficient agent memory mechanisms for longitudinal clinical care?} by analyzing the performance, efficiency, component
contributions and generalization of MedCache.

\subsection{MedCache Improves the Accuracy--Context Trade-off}
\label{sec:main-results}

Fig.~\ref{fig:accuracy-context} compares MedCache with the centralized
single-agent and medical multi-agent systems evaluated in
Tab.~\ref{tab:existing_memory_results}. MedCache achieves the highest overall
accuracy at .936. The strongest existing baseline, ClinicalAgents, reaches
.915 while processing 31.5k input tokens per question. MedCache numerically
improves accuracy by .021 while using 8.7k tokens, a 3.6$\times$ reduction. Overall,
MedCache lies on the accuracy--context Pareto frontier, providing the strongest
accuracy while maintaining a compact per-query context.

\subsection{Robustness to Increasing Distractors}
\label{sec:validity-results}

We hold each query and gold answer fixed while increasing the amount of stale and off-specialty evidence. \emph{Clean} uses the original record. \emph{Moderate} adds eight stale laboratory values and 40 off-specialty facts. \emph{Heavy} increases these counts to 20 and 100. All stale values are verified so that they do not affect the gold answer.

\begin{figure}[t]
\centering
\includegraphics[width=\columnwidth]{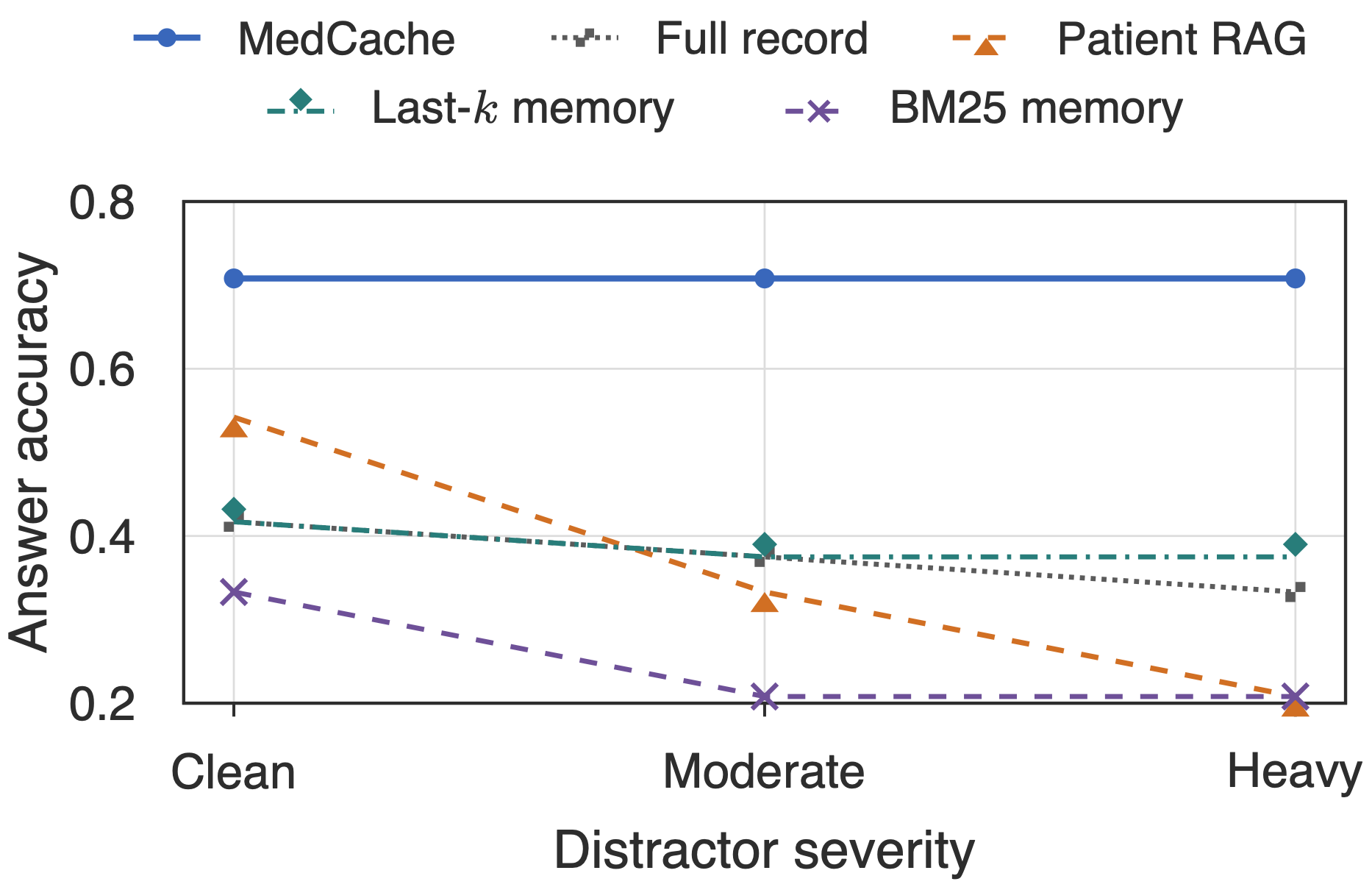}
\caption{Answer accuracy under increasing distractor severity.}
\label{fig:distractors_tiers}
\end{figure}

Fig.~\ref{fig:distractors_tiers} reports that MedCache maintains .708 accuracy from Clean through Heavy. RAG falls from .542 to .208, BM25 memory from .333 to .208, and the full record from .417 to .333. Last-$k$ memory reaches .417 on Clean and .375 on Heavy. At the curation stage, our validity-window policy retains the gold fact on every question at every severity (1.000), matching the oracle. The result identifies temporal validity as the mechanism that prevents stale evidence from entering the answering context.

\subsection{Ablation Analysis}
\label{sec:ablation-results}

Tab.~\ref{tab:component-ablation} reports a fixed-seed rerun that removes one component at a time from the full pipeline. Results are reported separately for long-context evidence retrieval, cross-time aggregation, and cross-specialty reasoning to show which type of question is affected by each component. Removing the learned gate causes the largest overall drop, mainly on evidence retrieval and cross-time aggregation. Validity curation is particularly important for cross-time questions, while specialty ownership provides smaller gains on both subsets. The Cross-Specialty improvement under two ablations corresponds to only two questions and does not offset the larger losses elsewhere.

\begin{table}[t]
\centering
\caption{Fixed-seed component ablation on 188 questions:
82 long-context evidence-retrieval questions, 46 cross-time
aggregation questions, and 60 cross-specialty reasoning questions.}
\label{tab:component-ablation}
\small
\setlength{\tabcolsep}{3.2pt}
\begin{tabular}{@{}lcccc@{}}
\toprule
Configuration & Overall & Evid & C-Time & C-Spec. \\
\midrule
\textbf{Full MedCache}
    & \textbf{.920} & .939 & .935 & .883 \\
$-$ learned gate
    & .766 & .707 & .674 & .917 \\
$-$ validity curation
    & .840 & .854 & .717 & .917 \\
$-$ specialty ownership
    & .888 & .890 & .891 & .883 \\
\bottomrule
\end{tabular}
\end{table}

\subsection{Serving Efficiency Under Expanding Context}
\label{sec:efficiency-results}

We next examine how serving efficiency changes as the available clinical
knowledge context expands. We increase the guideline registry from  8 to 24
modules, representing progressively broader specialty coverage and additional
guideline knowledge. A centralized system must manage an increasingly large
shared guideline context. MedCache instead selects the relevant specialist
module and combines its stable guideline prefix with a dynamic, routed
patient-evidence suffix. 

\begin{table}[t]
\centering
\caption{Serving efficiency as the available guideline context expands.
\emph{Modules} is the number of guideline modules in the registry.
\emph{Reuse} is the fraction of prefix tokens served from cache, and TTFT is
the median over 60 requests per configuration and three seeds.}
\label{tab:prefix-scaling}
\small
\setlength{\tabcolsep}{3.8pt}
\renewcommand{\arraystretch}{1.05}
\begin{tabular}{@{}ccccc@{}}
\toprule
Guideline & \multicolumn{2}{c}{Prefix reuse} &
             \multicolumn{2}{c}{TTFT (s)} \\
\cmidrule(lr){2-3}
\cmidrule(l){4-5}
modules & Central & MedCache & Central & MedCache \\
\midrule
8  & .888 & .935 & .212 & .164 \\
16 & .399 & .935 & .472 & .164 \\
24 & .248 & .926 & .716 & .165 \\
\bottomrule
\end{tabular}
\end{table}

As the available guideline context expands from 8 to 24 modules, centralized
prefix reuse falls from .888 to .248 and median TTFT increases from .212 to
.716\,s. MedCache remains nearly unchanged: prefix reuse stays above .926 and
median TTFT remains near .165\,s. Consequently, its TTFT advantage increases
from 1.29$\times$ with 8 modules to 4.34$\times$ with 24 modules. Selective
specialist activation therefore becomes increasingly valuable as the clinical
knowledge base grows: centralized execution repeatedly processes an expanding
shared guideline context, whereas MedCache activates only the guideline module
required by the current query.

\subsection{Generalization Across Backbones and Datasets}
\label{sec:generalization-results}

Tab.~\ref{tab:cross-model} compares all-active-memory access with owner-routed MedCache across two backbones. MedCache improves Qwen accuracy from .904 to .936 while reducing input context from 11.0k to 8.7k tokens. On Mistral, both systems reach .920 accuracy, but MedCache uses 8.4k rather than 10.7k tokens. Thus, owner-aware routing consistently reduces context, while its accuracy gain depends on the backbone. Across three gate cross-validation seeds, MedCache reaches $.919 \pm .007$ overall accuracy.

\begin{table}[t]
\centering
\caption{Cross-backbone accuracy and mean input tokens.}
\label{tab:cross-model}
\small
\setlength{\tabcolsep}{4.2pt}
\renewcommand{\arraystretch}{1.05}
\begin{tabular}{@{}lcccc@{}}
\toprule
& \multicolumn{2}{c}{Qwen 3.6-35B-A3B} &
\multicolumn{2}{c}{Mistral-24B} \\
\cmidrule(lr){2-3}
\cmidrule(l){4-5}
System & Acc. & In tok. & Acc. & In tok. \\
\midrule
All active memories
& .904 & 11.0k
& \textbf{.920} & 10.7k \\
\textbf{MedCache}
& \textbf{.936} & \textbf{8.7k}
& \textbf{.920} & \textbf{8.4k} \\
\bottomrule
\end{tabular}
\end{table}

The same routing behavior transfers to eICU-CRD 2.0~\cite{pollard2018eicu}. On 360 questions from 150 ICU stays, the learned gate raises cross-owner accuracy from .583 to .858 and achieves .856 overall while reading 27.5\% of patient memory. Selecting every active specialty reaches a similar .853 accuracy but reads 60.5\% of memory. External note-based tests show a similar advantage under irrelevant context: after adding distractors, MedCache changes from .632 to .640 on EHRNoteQA~\cite{kweon2024ehrnoteqa} and remains at .440 on LongHealth~\cite{adams2024longhealth}, whereas full-context accuracy falls from .752 to .572 and from .823 to .247, respectively. Full context remains stronger on the clean note datasets, indicating that MedCache primarily generalizes to settings with stale, irrelevant, or specialty-distributed evidence. Detailed results on external datasets can be found in the Appendix.

\section{Limitations}
\label{sec:discussion-limitations}

LEMMA-Bench contains 91 patients and 188 multiple-choice questions from a single retrospective MIMIC-IV cohort. Although the records span long clinical histories, broader validation requires multi-institutional data, open-ended clinical decisions, prospective workflows, and clinician-reviewed outcomes.

\section{Conclusion}
We introduced LEMMA-Bench to study memory design for longitudinal clinical agents. Our findings show that effective systems must maintain a valid patient state, organize evidence into focused but overlapping specialty memories, retrieve from the appropriate owners, and coordinate specialists only when needed. Guided by these findings, our proposed memory framework MedCache achieves the highest accuracy among evaluated systems while maintaining high computational efficiency.

\bibliography{aaai2027}

\clearpage

\section{Appendix}

\subsection{Additional Results}

\paragraph{Owner-routing transfer to eICU.}
We construct a cross-dataset evidence owner-routing evaluation test from eICU-CRD
2.0~\cite{pollard2018eicu}. It contains 360 current-value questions from 150
ICU stays: 170 same-owner and 190 cross-owner. Incorrect options are other
values of the same laboratory concept from the same stay. The owner gate uses
frozen Qwen embeddings and patient-grouped five-fold cross-validation, with
all questions from one stay kept in the same fold. The ten-specialty registry
is unchanged from LEMMA-Bench.

Table~\ref{tab:app-eicu-routing} gives the complete selector breakdown. The
learned gate reaches .856 overall and .858 on cross-owner questions, compared
with .433 and .063 for requester-only selection. BM25 improves cross-owner
accuracy to .342 but recovers only 70.3\% of the registered owners. The learned
gate recovers every registered owner and matches the active-specialty union on
cross-owner accuracy, while selecting 1.53 rather than 4.75 specialties and
reading 27.5\% rather than 60.5\% of patient memory. Central full memory reads
the entire record and reaches .842 overall, below both selective configurations.

\begin{table*}[t]
\centering
\caption{Owner selection on the eICU-CRD transfer set. Owner found is owner recall; Spec.\ is the mean number of selected specialties; Mem.\ is the fraction of patient memory read.}
\label{tab:app-eicu-routing}
\small
\setlength{\tabcolsep}{5.0pt}
\begin{tabular}{@{}lcccccc@{}}
\toprule
Selector & Overall & Same & Cross & Owner found & Spec. & Mem. \\
\midrule
Requesting specialty only & .433 & .847 & .063 & .472 & 1.00 & .160 \\
BM25 owner selector       & .583 & .853 & .342 & .703 & 1.68 & .241 \\
\textbf{Learned owner gate}
                          & \textbf{.856} & .853 & \textbf{.858}
                          & \textbf{1.000} & 1.53 & .275 \\
Active-specialty union    & .853 & .847 & .858 & 1.000 & 4.75 & .605 \\
Central full memory       & .842 & .841 & .842 & 1.000 & 10.00 & 1.000 \\
\bottomrule
\end{tabular}
\end{table*}

\paragraph{Patient-scoped longitudinal-note controls.}
EHRNoteQA~\cite{kweon2024ehrnoteqa} and
LongHealth~\cite{adams2024longhealth} test a different failure mode: documents
from other patients are inserted as distractors. Because such documents should
not be visible in a patient-specific chart, we additionally evaluate a full-
context control that filters by patient identity before answering. The clean
and distractor conditions use the same target questions and answers.

\begin{table*}[t]
\centering
\caption{Accuracy and mean input tokens on patient-scoped longitudinal-note
controls. Distr.\ adds documents from other patients before patient-identity
filtering.}
\label{tab:app-note-scope}
\small
\setlength{\tabcolsep}{5.0pt}
\begin{tabular}{@{}lclcccc@{}}
\toprule
& & & \multicolumn{2}{c}{Clean} & \multicolumn{2}{c}{Distr.} \\
\cmidrule(lr){4-5}
\cmidrule(l){6-7}
Dataset & $n$ & System & Acc. & In tok. & Acc. & In tok. \\
\midrule
EHRNoteQA & 250 & Scoped full context
          & \textbf{.760} & 2.3k & \textbf{.740} & 2.3k \\
          &     & MedCache
          & .632 & \textbf{2.1k} & .640 & \textbf{2.1k} \\
\midrule
LongHealth & 400 & Scoped full context
           & \textbf{.825} & 11.9k & \textbf{.820} & 11.9k \\
           &     & MedCache
           & .440 & \textbf{4.3k} & .440 & \textbf{4.3k} \\
\bottomrule
\end{tabular}
\end{table*}

Table~\ref{tab:app-note-scope} shows that patient-scoped full context is stable
after distractors are added: accuracy changes from .760 to .740 on EHRNoteQA
and from .825 to .820 on LongHealth. MedCache changes from .632 to .640 and
remains at .440, respectively, while using less context. Full context therefore
remains more accurate on these note benchmarks once patient identity is
controlled. These results show that MedCache is robust to irrelevant documents,
but they do not establish an accuracy advantage from specialty routing in a
setting where simple patient-identity filtering removes the distractors.

\subsection{LEMMA-Bench Details}
\paragraph{Cohort and unit of evaluation.}
LEMMA-Bench contains 188 multiple-choice questions from 91 patients in MIMIC-IV v3.1. Patients were selected to have long histories. Every patient has at least eight hospital admissions, and the cohort contains 1,152 admissions in total, a mean of 12.7 per patient with a median of 11 and a range of 8 to 39. The average record spans 5.8 years, with a median of 5.1 years and a longest record of 13.9 years. Across these histories, the benchmark covers 193,394 laboratory measurements, a mean of 2,125 per patient and a median of 1,324, drawn from 381 distinct laboratory concepts, of which an average patient has 88. The dataset statistics can be found in Tab.~\ref{tab:app-lemma-cohort}.

\begin{table*}[t]
\centering
\caption{LEMMA-Bench cohort statistics. Patient-level summary statistics are computed over the 91 patients that contribute at least one surviving benchmark question.}
\label{tab:app-lemma-cohort}
\small
\setlength{\tabcolsep}{8.0pt}
\begin{tabular}{@{}lrrrrr@{}}
\toprule
Quantity & Total & Mean & Median & Min & Max \\
\midrule
Patients                     & 91      & --    & --    & --  & -- \\
Hospital admissions          & 1,152   & 12.7  & 11    & 8   & 39 \\
Record span (years)          & --      & 5.8   & 5.1   & 0.1 & 13.9 \\
Laboratory measurements      & 193,394 & 2,125 & 1,324 & 33  & 8,994 \\
Distinct laboratory concepts & 381     & 88    & 84    & 33  & 171 \\
Benchmark questions          & 188     & 2.1   & 2     & 1   & 4 \\
\bottomrule
\end{tabular}
\end{table*}

The unit of evaluation is a question at a specified record cutoff: only information available at or before that cutoff may be used. The benchmark construction procedure is described in main paper.

\paragraph{Question record.}
Each released item stores the patient identifier, question, options, gold option, record cutoff, task label, required specialties, supporting clinical-atom identifiers, and supporting guideline-clause identifiers. The support fields are used for construction filters and evaluation only. They are never supplied to the router, retriever, specialists, or coordinator at test time. Options are shuffled after filtering and the gold index is recomputed from answer text.

\begin{table}[t]
\centering
\caption{LEMMA-Bench composition. Cross-specialty takes precedence when a question also aggregates evidence over time.}
\label{tab:app-lemma-composition}
\scriptsize
\setlength{\tabcolsep}{3.0pt}
\begin{tabular}{@{}lcl@{}}
\toprule
Question type & Count & Required operation \\
\midrule
Evidence retrieval & 82 & Locate one relevant state \\
Cross-time aggregation & 46 & Combine recurring measurements \\
Cross-specialty reasoning & 60 & Combine specialty constraints \\
\midrule
Total & 188 & \\
\bottomrule
\end{tabular}
\end{table}

\paragraph{Task labels.}
The labels in Table~\ref{tab:app-lemma-composition} are assigned from recorded support rather than question wording. A question is cross-time when its support contains at least two values of the same concept at different times. It is cross-specialty when the support spans multiple specialties; this label takes precedence. The first two groups together form the 128 evidence questions. Among them, 85 are same-owner and 43 are cross-owner relative to the requesting specialty.

\paragraph{Metrics and uncertainty.}
Answer accuracy is exact multiple-choice accuracy after a common output parser. We report Overall and the three task-specific accuracies above. Routing experiments additionally report owner recall, exact selected-set accuracy, the mean number of selected specialties, and the fraction of patient memory read. The 95\% confidence intervals in the main results use 5,000 cluster bootstrap replicates over patients, so all questions from one patient are resampled together.

\subsection{Method Details}

\paragraph{Overlapping ownership.}
Let $\mathcal{O}(c)$ be the fixed registry mapping a clinical concept to one or more specialties. Specialty $s$ stores
\[
 M_{p,s}(t_q)=
 \{a_i\in V_p(t_q):s\in\mathcal{O}(c_i)\}.
\]
An atom assigned to several specialties keeps one source identifier across all partitions, allowing the routed union to be deduplicated without losing provenance. Patient updates recompute affected validity intervals, trajectories, partitions, and active-specialty metadata.

\paragraph{Owner gate and read set.}
The gate embeds the question with a frozen Qwen encoder and applies a two-layer classifier trained with patient-grouped cross-validation. For an evidence question with requester $r$ and predicted owner $\hat{o}$, the read set is
\[
 R(q)=M_{p,r}(t_q)\cup M_{p,\hat{o}}(t_q).
\]
The assembler deduplicates shared atoms, preserves complete required trajectories, ranks the remaining facts against the question and selected guideline, and applies the token budget only after these constraints.

Because the owner targets and memory placement use the same concept-to-specialty registry, owner recall measures \emph{registry-consistent routing}, not an independent judgment of clinical sufficiency. If the registry is complete and correct, selecting the registered owner is operationally sufficient for the configured memory system. It still does not test whether the registry itself omitted or misassigned clinically relevant evidence.

\paragraph{Adaptive execution.}
Evidence-retrieval and cross-time questions use one requesting specialist over the routed union, even when that union contains facts from several specialty memories. Only a cross-specialty decision activates the relevant specialists concurrently. Each specialist reads its own partition and guideline module and emits a structured assessment containing its decision, evidence, uncertainty, conflicts, and source identifiers. The coordinator then receives these bounded assessments, aligns shared evidence, and produces one answer. Thus the coordinator is conditional, not an always-on stage.

\paragraph{Versioned guideline context and caching.}
Guideline modules are immutable within a version and record their topic, version, applicability metadata, provenance, content hash, and token length. The selected module is serialized as the stable system-prefix; routed patient evidence and the question form the dynamic suffix. vLLM automatic prefix caching reuses engine-managed state only when the prefix bytes and version match. Updating a guideline creates a new hash and therefore a new cache key rather than mutating an existing prefix.

\subsection{Baseline Details}
We evaluate SAS memory systems including raw long-context baselines, RAG~\cite{lewis2020retrieval}, Zep~\cite{rasmussen2025zep}, MemGPT(Letta)~\cite{packer2023memgpt}, Mem0~\cite{chhikara2025mem0}, and MAS memory systems including MDAgents~\cite{kim2024mdagents}, Med-SRAF~\cite{li2026medsraf}, ClinicalAgents~\cite{ge2026clinicalagents}. 

\subsubsection{Common Evaluation Harness}

All baselines are evaluated through a common harness so that differences primarily reflect their memory and reasoning designs. Each system begins from the same normalized, chronologically ordered patient record and then applies its own memory-construction, retention, partitioning, or retrieval policy. All systems use the same multiple-choice output instruction and have access to the same underlying clinical-guideline sources. When a method consumes compiled guideline modules directly, it receives the same module content used by MedCache; methods designed to retrieve from source documents instead construct their guideline context through their native retrieval procedure.

Query-time generation uses Qwen3.6-35B-A3B served through vLLM in the main experiments and Mistral-Small-24B in the cross-backbone replication. All generations use greedy decoding with temperature $0$. 

A shared deterministic parser extracts the selected option from outputs such as
\texttt{ANSWER: A}, \texttt{ANSWER: (A)}, and \texttt{**A**}. All reported experiments were rerun using the final parser after correcting an earlier failure to recognize some chain-of-thought output formats.

We record the number of model calls, prompt tokens, completion tokens, and wall-clock latency for each evaluated item. For multi-agent methods, costs are summed across all routing, recruitment, specialist, coordination, and verification calls invoked by the system.

\subsubsection{Multi-Agent Baselines}

\paragraph{MDAgents.}
MDAgents~\cite{kim2024mdagents} adaptively selects a reasoning configuration according to estimated query complexity. An initial LLM call classifies each query as basic, intermediate, or advanced. A basic query is then answered by a single agent. For an intermediate query, the system recruits three experts, obtains an independent assessment from each expert, and uses a moderator to synthesize their assessments. The advanced configuration recruits two team leads, obtains an initial multidisciplinary assessment, and passes that assessment for final review.

Patient memory is shared and unpartitioned, and all answering agents receive the same patient record. Applicable guideline modules are supplied according to the clinical criterion addressed by the question rather than according to gold patient-evidence owners.

To make execution deterministic, the intermediate configuration always recruits three experts, and the advanced configuration always recruits two team leads. The difficulty-classification call has a maximum output length of 20 tokens. Excluding this initial classification call, the basic, intermediate, and advanced execution stages use one, five, and three model calls, respectively. Thus, the complete paths contain two, six, and four calls when difficulty classification is included.

For the intermediate path, the recruitment call has an 80-token output budget, each expert has a 900-token output budget, and the moderator has a 1,500-token output budget. For the advanced path, recruitment has a 60-token output budget, the initial assessment has an 800-token output budget, and final review has a 1,500-token output budget. 

\paragraph{Med-SRAF.}
Med-SRAF~\cite{li2026medsraf} routes each query to relevant knowledge dimensions, retrieves evidence independently within each selected dimension, and reconciles the resulting assessments through a fusion agent. The original system uses a UMLS-based router. Because the required UMLS resources were not included in the released implementation and require separate licensed access, we replace this component with an LLM router that selects at most three specialty dimensions from the question text.

Patient memory is partitioned by specialty, and each dimension agent can retrieve only from its own partition. Evidence is selected using BM25 under an 800-token evidence budget. Each selected dimension invokes one specialist with an 800-token output budget. The specialist returns a specialty-specific assessment and a provisional answer.

The router has a 40-token output budget. If it selects $m$ specialty dimensions, the complete execution requires $m+2$ calls: one routing call, $m$ specialist calls, and one fusion call. Routing latency is measured separately from specialist and fusion generation latency so that it can be compared with the MedCache owner gate.

\paragraph{ClinicalAgents.}
ClinicalAgents~\cite{ge2026clinicalagents} combines a Working Memory containing current patient information with an Experience Memory containing general knowledge and related historical cases. It reasons through a perceive--hypothesize--verify--update procedure in which an initial hypothesis can be revised when the supporting evidence is incomplete or inconsistent. 

Working Memory is instantiated as the shared, unpartitioned patient record. Experience Memory contains the applicable guideline modules and the two most similar historical cases (no overlap with benchmark cases) retrieved using BM25. 

The system uses two model calls, each with a maximum output length of 2,000 tokens. The first call proposes a hypothesis, identifies its supporting evidence, and lists any missing clinical concepts. The second call re-examines the record and guidelines, revises the hypothesis when necessary, and commits to a final answer. The final 600 characters of the first response are passed to the second call, and each retrieved historical case is truncated to 600 characters. Backtracking is limited to one revision round.

\subsubsection{Memory-Policy Baselines}

We implement MemGPT, Zep, and Mem0 as policy-level baselines behind a common interface that determines which patient facts are provided to the query-time reader. Each policy operates under a budget of 20 memory entries. Long-Context instead provides the complete patient timeline and therefore does not use the bounded-memory interface.

\paragraph{MemGPT.}
The MemGPT policy~\cite{packer2023memgpt} maintains a query-independent working memory containing the K most recent patient facts and one summary block covering all earlier facts. The summary is generated once per patient and remains fixed across all questions for that patient. The resulting working set therefore contains at most 20 entries.

\paragraph{Zep.}
The Zep policy~\cite{rasmussen2025zep} retains the complete patient fact collection and performs retrieval at query time rather than explicitly evicting older facts. For each question, BM25 ranks all available patient facts, and the top 20 are passed to the reader.

\paragraph{Mem0.}
The Mem0 policy~\cite{chhikara2025mem0} uses add-only storage and combines relevance with recency during retrieval.

\paragraph{Long-Context.}
The Long-Context baseline places the complete chronologically ordered patient record in the prompt and does not impose a fact budget. It provides an unbounded-context reference for evaluating the effects of memory selection and compression.

\end{document}